\documentclass[lettersize,journal]{IEEEtran}
\usepackage{amsmath,amsfonts}
\usepackage{algorithmic}
\usepackage{algorithm}
\usepackage{array}
\usepackage[caption=false,font=normalsize,labelfont=sf,textfont=sf]{subfig}
\usepackage{textcomp}
\usepackage{stfloats}
\usepackage{url}
\usepackage{multirow}
\usepackage{verbatim}
\usepackage{makecell}
\usepackage{graphicx}
\usepackage{hhline}
\usepackage{xcolor}
\usepackage{cite}
\usepackage{bm}
\usepackage{booktabs}
\usepackage{amsmath}
\usepackage{upgreek}
\usepackage{float}
\begin{document}

\title{Inter and Intra Prior Learning-based Hyperspectral Image Reconstruction Using Snapshot SWIR Metasurface}

\author{Linqiang Li, Jinglei Hao, Yongqiang Zhao, Pan Liu, Haofang Yan, Ziqin Zhang,~\IEEEmembership{Member,~IEEE,} and Seong G. Kong,~\IEEEmembership{Senior Member,~IEEE}


\thanks{This work was supported by the National Natural Science Foundation of China (NSFC) (No. 61771391), the Key R$\&$D plan of Shaanxi Province (No.
2020ZDLGY07-11).\par
L. Li, Y. Zhao, P. Liu, F. Yan and Z. Zhang are with the School of
Automation, Northwestern Polytechnical University, Xi’an 710129,
China (e-mail: lilinqiang0921@mail.nwpu.edu.cn; zhaoyq@nwpu.edu.cn; liu$\_$pan@mail.nwpu.edu.cn; yanhaofang@mail.nwpu.edu.cn; zhangziqin@mail.nwpu.edu.cn).\par
J. Hao is with the college of information and control engineering, Xi'an university of architecture and technology, Xi’an 710311, China (e-mail: orangeshirley1991@163.com).\par
S. G. Kong is with Department of Computer Engineering, Sejong University, Seoul 05006, Korea (e-mail:skong@sejong.edu).}}



\maketitle

\begin{abstract}

Shortwave-infrared(SWIR) spectral information, ranging from 1 $\bm{\upmu}$m to 2.5$\bm{\upmu}$m, overcomes the limitations of traditional color cameras in acquiring scene information. However, conventional SWIR hyperspectral imaging systems face challenges due to their bulky setups and low acquisition speeds. This work introduces a snapshot SWIR hyperspectral imaging system based on a metasurface filter and a corresponding filter selection method to achieve the lowest correlation coefficient among these filters. This system offers the advantages of compact size and snapshot imaging. We propose a novel inter and intra prior learning unfolding framework to achieve high-quality SWIR hyperspectral image reconstruction, which bridges the gap between prior learning and cross-stage information interaction. Additionally, We design an adaptive feature transfer mechanism to adaptively transfer the contextual correlation of multi-scale encoder features to prevent detailed information loss in the decoder. Experiment results demonstrate that our method can reconstruct hyperspectral images with high speed and superior performance over existing methods. 
\end{abstract}


\section{Introduction}
\label{sec:Introduction}
\begin{figure}
    \centering
    \includegraphics[width=1.0\linewidth]{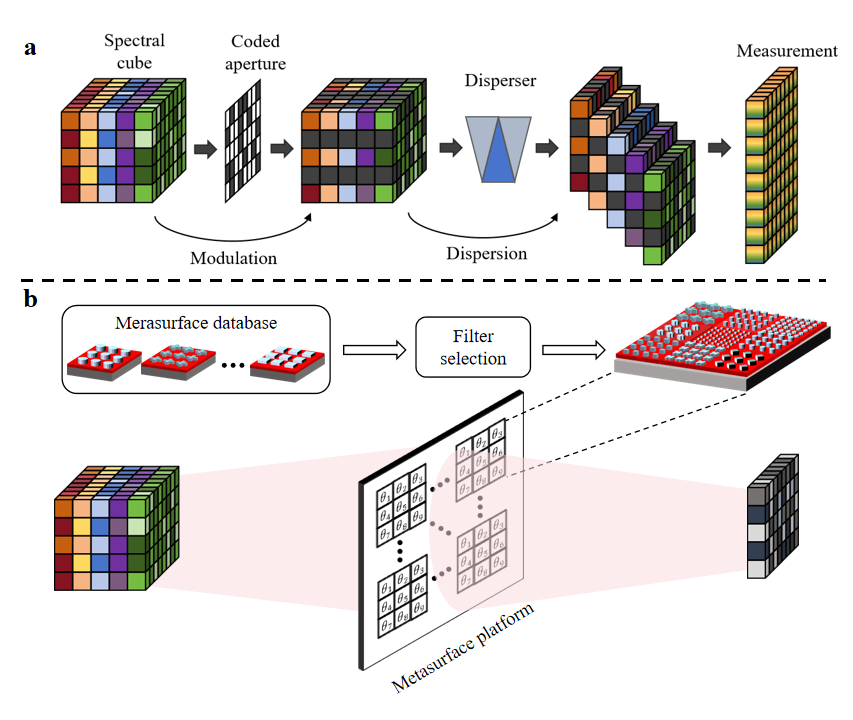}
    \caption{Schematic representation of different snapshot compressive imaging systems. (a) The architecture of the CASSI system; (b) Our snapshot SWIR hyperspectral imaging system based on a metasurface filter.}
    \label{fig1}
\end{figure}

Hyperspectral imaging (HSI), characterized by spatial-spectral data-cubes, offers rich spectral information beneficial in various fields such as medical diagnosis\cite{medical_1,medical_2}, food safety\cite{food_safety}, and remote sensing\cite{Remote_sensing_1, Remote_sensing_2}. Conventional hyperspectral imaging systems typically involve sequential temporal scanning of either the spatial\cite{spatial_1, spatial_2} or spectral\cite{spectral} domain. However, such scanning procedures result in slow imaging processes that are unsuitable for dynamic or real-time applications. Recently, Snapshot Compressive Imaging (SCI) systems like coded aperture snapshot spectral imaging (CASSI)\cite{hdnet, CST, MST} and broadband filter imaging\cite{broadband_filter_imaging1,broadband_filter_imaging2} provide an elegant solution. These systems utilize a 2D detector to capture a 3D hyperspectral cube in a single and reconstruct the HSI based on compressed sensing principles. As shown Fig. \ref{fig1}(a), CASSI spatially modulates the 3D HSI with a coded aperture and spectrally shifts it with dispersive elements. The data captured by the sensor is an aliasing of different monochromatic images. However, the use of complex optical components has made these systems relatively large and not easily portable or integrated with mobile devices.\par

Unlike CASSI, broadband filter imaging systems directly modulate spectral information through platform composed of basic filter patterns integrated with sensors as shown in Fig. \ref{fig1}(b). In such a architecture, the layout of apertures and disperses  is no longer required. Xiong \textit{et al.}\cite{dynamic_brain} first proposed a silicon real-time spectral imaging chip in visible band. He \textit{et al.}\cite{he2024meta} extended metasurface-based snapshot imaging systems into the near infrared (NIR) range of 0.7-1$\bm{\upmu}$m, and designed a meta attention network to achieve HSI reconstruction. However, these studies neglect spectral imaging in the shortwave infrared (SWIR) range of 1-2.5$\bm{\upmu}$m, which is significant in gas detection, standoff imaging, and material detection. \par

Additionally, the reconstruction quality of SCI systems heavily depends on filter selection and reconstruction algorithms. Existing methods have shortcomings in two aspects. The randomness of basic filter array is crucial for spectral coding efficiency and meeting the conditions for compressed information reconstruction. However, exhaustive search methods and evolutionary optimization methods\cite{filter_selection} require multiple spectral reconstructions to  verify the effectiveness of selected filters, making them extremely slow. Correlation-based methods\cite{dynamic_brain,he2024meta} lack efficient optimization processes for selecting optimal filter array with the lowest correlation. \par

Existing algorithms for HSI reconstruction can be categorized into three main types: traditional model-based approaches, deep learning-based approaches and deep unfolding network methods. Traditional model-based methods\cite{model1, model2, model3, model4} address an ill-posed optimization problem iteratively using manually designed prior. While these methods offer high interpretability, they are constrained by the limitations of hand-crafted priors and exhibit slow reconstruction speeds. Deep learning-based methods\cite{meng2020end, learn1, learn2, learn3} leverage the nonlinear mapping ability of deep networks to directly learn a mapping from 2D measurements to 3D hyperspectral cubes, leading to significant advancements in reconstruction quality and speed. However, they lack explicit representation of the imaging mechanism and process HSI reconstruction as a black-box, which limits further improvement. Deep unfolding networks (DUNs)\cite{DGSMP,DNUs1, DNUs2, ADMM-Net, DNUs3, DNUs4, DNUs5} address the challenge of limited interpretability in black-box networks by combining deep learning with mathematical models. They execute the iterative process using a gradient descent module and enhance the intermediate output through a deep prior module. Although existing DUNs address certain limitations of model-based and learning-based methods, several challenges remain. Firstly, several works utilize deep prior modules to learn HSI's implicit priors to guide spectral reconstruction, such as low rank\cite{low_rank_prior}, Gaussian mixture models\cite{gaussian_prior},  and deep subspace projection \cite{deep_subspace_projection_prior}. However, the knowledge acquired by each individual prior module cannot be effectively shared throughout the multi-stage optimization process, resulting in sub-optimal prior learning. Secondly, skip feature interaction between the encoder and decoder of U-Net can effectively decrease information loss caused by up- and down-sampling operations. But existing works use simple skip residual concatenation\cite{DNUs2, skip1}, which cannot complement detailed features of HSI, while rough multi-scale feature fusion\cite{DNUs4} may introduce additional noise. \par
In this paper, we propose a novel filter selection method for simulating a metasurface-based snapshot spectral imaging system SWIR range of 1-2.5$\bm{\upmu}$m and a corresponding deep unfolding network-based algorithm to achieve HSI reconstruction. For the imaging system simulation, we consider sufficient spectral variety and high transmission, designing thousands of metasurface units constructed by silicon nanopillars. Then we obtain the optimal basic filter array with our proposed optimization schedule, ensuring the lowest mutual correlation between units. Comparative experiments demonstrate that our proposed filter selection method significantly benefits reconstruction quality. In terms of the reconstruction algorithm, we propose a novel inter and intra prior learning network with high performance. To achieve global prior learning in deep unfolding network, we introduce a query prior to interact with each stages and learn HSI's low rank prior. Considering the information loss between encoder and decoder, we design an adaptive feature transfer module to construct contextual correlation features of different scales and transfer them into the decoder. \par
The contributions of this paper can be summarized as follows:
\begin{itemize}
\setlength{\itemsep}{5pt}

    \item Proposes a novel filter selection method for sufficient coding of incident spectrum and satisfactory compressed information reconstruction to construct a metasurface-based SWIR snapshot spectral imaging system.
    \item Proposes a novel inter-stage prior learning and intra-stage feature transfer deep unfolding network (ERRA) for HSI reconstruction.
    \item Introduces an elegant cross-stage low-rank prior learning module to learn HSI’s low-rank prior across stages.
    \item Introduces an efficient adaptive feature transfer module to  effectively transfer  contextual  correlation knowledge from the encoder to the decoder.
    
\end{itemize}

The rest of paper is organized as follows. Section \ref{sec:Related work} reviews related work. Section \ref{sec:METHOD} describes the details of the proposed filter selection method and HSI reconstruction algorithm. In Section \ref{sec:Experiments}, we conduct comprehensive experiments and ablation studies. Section \ref{sec:Conclusion} presents the conclusion. 
\section{Related work}
\label{sec:Related work}
\textbf{Filter selection} Arad and Ben-Shahar\cite{filter_selection} first identified that the accuracy of HSI reconstruction was heavily dependent on the selection of filter array and proposed an evolutionary optimization to select the optimal filter array. However, it requires training the HSI reconstruction network multiple times. Cui \textit{et. al} \cite{dynamic_brain} utilize the inner product of filter spectral response (FSR) to measure mutual correlation in the array, They set initial units from the entire metasurface database and optimize the array by replacing units that exceed the correlation upper threshold. Although this avoids redundant training processes, it still requires multiple iterations and the selected filter has poor randomness. He \textit{et. al} \cite{he2024meta} use the correlation coefficient as a measurement and strive to minimize the average correlation in the whole array, but it still lack a simple and effective optimization procedure, resulting in non-uniformity of mutual correlation coefficients in the array. 

\textbf{Model-based Methods.} Traditional methods solve the inverse problem of spectral reconstruction by modeling a system of linear equations. The target spectrum can be effectively reconstructed by utilizing prior knowledge as regularization terms such as total variation\cite{TV}, dictionary learning\cite{dic1,dic2}, non-local low rank\cite{non_local1,non_local2}, and Gaussian mixture models\cite{guaiss}. In several studies, sparse optimization-based methods, which rely on the assumption of HSI's sparsity as prior and use $l1$ to regularize the solution, have demonstrated better performance in reconstruction quality. In \cite{dic_1}, a Gaussian kernel-based sparse transform is proposed to solve reconstruction problem of specific mapping. \cite{dic_2} performs over-complete dictionary learning on HSI, then solves least square regression with sparse regular terms. This combination of dictionary learning and sparsity regularization has been widely used in HSI imaging of broadband filter arrays. \par
However, these method cannot bridge the gap between hand-crafted prior and real-world spectra, resulting in instability in the reconstruction results. The iterative process takes too long,  making it difficult to meet the needs of real-time imaging.\par
\textbf{Deep Learning-based Methods.} Depending on the representation ability of neural networks, deep learning-based HSI reconstruction methods have attracted widespread attention in recent years. The pioneering  method \cite{robles2015single} used convolutional neural networks to extract the spectral correlation between local pixels for spectral reconstruction. Stiebel \textit{et al.}\cite{stiebel2018reconstructing} introduced U-Net\cite{ronneberger2015u} to further mine non-local similarity and multi-scale detailed features. Subsequently, many variants of convolution-based spectral reconstruction models were derived\cite{can2018efficient,zhao2020hierarchical}. Later, TSA-Net \cite{meng2020end}stacked spatial-spectral transformers to reconstruct HSI. BIRNAT\cite{BIRNAT} integrated the expressive power of an end-to-end convolutional framework with the sequence correlation extraction capability of bidirectional Recurrent Neural Networks (RNNs) for HSI reconstruction. To further improve the efficiency of Tranformer for compressive sensing problems, CST\cite{CST} and MST\cite{MST} were proposed to reform the attention mechanism to reveal the intrinsic characteristics of HSI and reduce computational and memory costs. Although high quality and real-time reconstruction have been achieved, these brute-force methods lack interpretability and full utilization of transmittance characteristics.\par

\textbf{Interpretable network Methods.} Commonly used optimization algorithms for HSI reconstruction, such as HQS\cite{HQS}, ADMM\cite{ADMM}, PGD\cite{PGD}, GAP\cite{GAP}, can be disentangled into data fidelity and regularization terms, leading to iterative optimization algorithms that alternately solve the data subproblem and the prior subproblem. Interpretable network methods usually solve the prior subproblem as a denoising subproblem while retaining data subproblem, combining the interpretability of model-based method and the strong generalization ability of deep learning-based methods. Plug-and-play methods \cite{PnP1, PnP2} leveraged pre-trained neural networks that have learned from extensive spectral data to enhance reconstruction accuracy in both spatial and spectral domains. Although such method shows strong generalization ability and are free from pre-training, they still face the challenge of being time-consuming due to the iterative gradient descent procedure. \par
Apart from Plug-and-play methods, deep unfolding networks unfold deep networks and insert data fidelity term into the training process, thus achieving joint optimization. For instance, ADMM-Net\cite{ADMM-Net} unfolded the alternating direction method of multipliers with a convolutional neural network. DSSP\cite{DSSP} extended the half-quadratic splitting (HQS) method and created a spatial-spectral deep prior to enhance data fidelity. DGSMP \cite{DGSMP} presented an unfolding model estimation framework that utilizes the learned Gaussian scale mixture prior to enhance the model's performance. RDLUF\cite{DNUs4} introduces interaction between different stages with a sequence feature learning formula (similar to GRU\cite{GRU}), Song \textit{et al.}\cite{song2023optimization} implemented feature interaction between different stages using a channel self-attention mechanism, and replaced the process of combining inertia and original gradient terms with an adaptive network. \par
Although the above methods attempt image prior learning or information interaction between different stages, they still separate the two problems, and never try to solve them in the same module. Additionally, existing methods neglect efficient feature transfer from encoder to decoder stages, which is crucial for retaining detailed features.\par

\begin{figure*}[tp]
    \centering
    \includegraphics[width=1\linewidth]{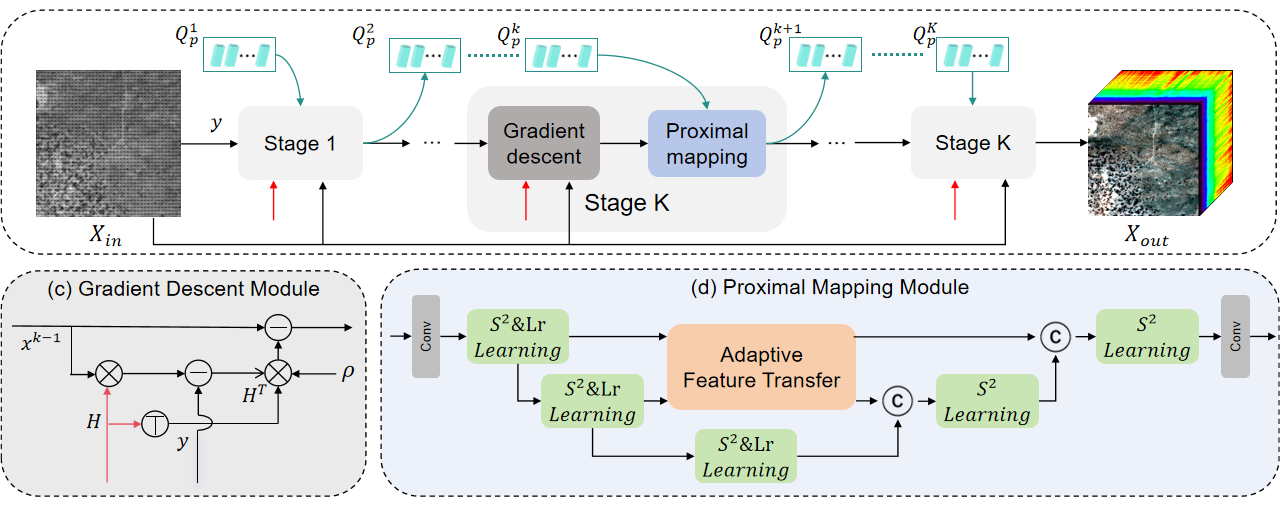}
    \caption{Illustration of the proposed ERRA for HSI reconstruction, Top: the overall architecture that consists of K stages, each of which which consists of a gradient descent module and a proximal mapping module. (a) Gradient descent module; (b) Proximal mapping module.}
    \label{fig2}
\end{figure*}

\section{Method}
\label{sec:METHOD}
\subsection{Problem description}
\label{subsec:Problem Description}
The metasurface-based SWIR snapshot spectral imaging system mainly consists of hardware encoding and spectra reconstruction. The hardware encoding encodes the incident spectrum through metasurface platform as presented in Fig. \ref{fig1}(b). Considering a HSI $X \in R ^ {\; H \times W \times \lambda}$, the captured HSI from target scenes is modulated via $\Theta$, and the measurement $Y \in R ^ {\; H \times W}$ on sensor plane is denoted as:
{
\setlength\abovedisplayskip{10pt}
\setlength\belowdisplayskip{10pt}
\begin{equation}
Y = \Theta \cdot X + n
\end{equation}
}%
where $\cdot$ denotes the element-wise multiplication, $n \in R^{\; H\times W}$ represents additive noise in the coding process. In Section \ref{subsec:Pattern optimization}, we describe the details of the hardware part and propose a new filter selection method for achieving optimal platform $\Theta$, which can achieve more efficient incident spectrum encoding and help to improve the subsequent spectral reconstruction performance.\par
The spectra reconstruction aims to recover high-quality image $X$ from its measurement $Y$, which is typically an ill-posed problem. Mathmatically, the solution of HSI reconstruction could be moduled as:
{
\setlength\abovedisplayskip{10pt}
\setlength\belowdisplayskip{10pt}
\begin{equation}
x = argmin \frac{1}{2} \Vert y-\Theta x \Vert _{2}^{2} +\lambda \phi(x)
\label{optimize}
\end{equation}
}%
the first term is data fidelity term, while the second term $\phi$ is regularization term. $\lambda$ denotes a regularization parameters.\par

\par
In the iterative shrinkage thresholding algorithm (ISTA), Eq. \ref{optimize} expressed as an iterative convergence problem through the following iterative function:
{
\setlength\abovedisplayskip{10pt}
\setlength\belowdisplayskip{10pt}
\begin{align}
    & r^{(k)} = x^{(k-1)}-\rho \Theta^\top (\Theta x^{(k-1)}-y),\label{gradient}  \\
    & x^{(k)} = argmin \frac{1}{2} \Vert x-r^{(k)} \Vert _{2}^{2} +\lambda \phi(x),\label{proximal}
\end{align}
}%
where $k$ denotes the number of ISTA iteration and $\rho$ denotes the step size. The Eq. \ref{gradient} is a gradient operation and Eq. \ref{proximal} can be solved by proximal mapping as following:
{
\setlength\abovedisplayskip{10pt}
\setlength\belowdisplayskip{10pt}
\begin{equation}
x^{(k)} = prox(r^{(k)})
\end{equation}
}

The ISTA algorithm iteratively updates $r^{(k)}$ and $x^{(k)}$ until convergence, but it mainly suffers from two problems. Firstly, the manually designed weight parameter has poor generalization ability, resulting in low fidelity of reconstructed HSI. Secondly, it requires pixel-by-pixel spectral reconstruction, thus the reconstruction time sharply increases with the increase in spatial resolution. To address these issues, we unfold the ISTA algorithm and integrate our designed proximal mapping network into the gradient descent step as detailed in Section \ref{ssec:Proximal mapping network}.
\subsection{Filter selection}
\label{subsec:Pattern optimization}

As shown in Fig. \ref{fig1}(b), the metasurface-based SWIR snapshot imaging system integrates an imaging plane onto a sensor plane. The imaging plane $\Theta \in R ^{\; H \times W \times \lambda}$ comprises a set of metasurface-based basic filter arrays made of $Si$ and fabricated on a $SiO_{2}$ substrate. These arrays consist of $3 \times 3$ different spectral filters obtained through our proposed filter selection method. Each spectral filter is constructed using the same metasurface units arranged in a regular layout.\par

Under such a design, each filter with random peaks captures a portion of spectral information in a local area (i.e, $3 \times 3$ pixels), while each pixel is also influenced by additional information from surrounding pixels. Additionally, we apply a compressed sensing-based algorithm in combination with deep learning to achieve HSI reconstruction, which requires optimizing the metasurface's transmission spectra according to compressed sensing principles. Based on the above two points, the filter selection must meet the criteria of minimizing coefficient correlation to effectively encode incident spectrum and adhere to the Restricted Isometry Property (RIP) for accurate compressed information reconstruction.\par
We design thousands of meta units and removal units with limited randomness under a constraint of transmittance curve gradient threshold. This pre-processed set of metasurface units is utilized as a metasurface dataset $M \in R^{\; N \times \lambda}$ for filter selection. Inspired by the farthest point sampling algorithm, we first calculate the coefficient correlation between units and select the metasurface with the lowest total correlation as a reference structure. Then, we search for units with the least correlation to the the reference structure within $M$ as a new structure. We repeat these steps until 9 units are selected, denoted as $\theta \in R^{\; 9 \times \lambda}$ as described in Algorithm \ref{alg:alg1}.
\begin{algorithm}[H]
\caption{Filter Selection}\label{alg:alg1}
\setlength{\baselineskip}{1.4\baselineskip}
\begin{algorithmic}[1]
\STATE {$\mathbf{Input: }$}$Entire \; Metasurface \; Dataset \; M \in R^{\; N \times \lambda}$
\STATE {$\mathbf{Output: }$}$Selected \; Metasurface \; \theta \in R^{\;9 \times \lambda}$
\STATE $\mathbf{Preprocess}$

\STATE $\mu_{i} = \frac{\sum_{k=1}^{\lambda}M[i,k]}{\lambda} \leftarrow mean \; of \; transmission \; spectra$
\STATE $\sigma_{i}^{2} = \frac{\sum_{k=1}^{\lambda}(M[i,k] - \mu_{i})^{2}}{\lambda} \leftarrow standard \; deviation$
\STATE $cov[i, j] = \frac{\sum_{k=1}^{\lambda}(M[i, k] - \mu_{i}) \times (M[j, k] - \mu_{j})}{\lambda}\leftarrow covariance$
\STATE $p[i, j] = \frac{cov[i,j]}{\sigma_{i} \times \sigma_{j}}\leftarrow pearson \; correlation \; coefficient$
\STATE $d \leftarrow initialized \ coefficient \ in \ R^{\; N}$
\STATE $index = min\_index(\frac{\sum_{j=1}^{N}|p[i,j]|}{N - 1})$ \hspace{1cm} \; $i\neq j$
\STATE $\theta[1] = index$

\FORALL {$i=2,\cdots,9$}
\STATE $\hspace{0.5cm} \; mask = p[index] > d$
\STATE $\hspace{0.5cm} \; d[mask] = p[index, mask]$
\STATE $\hspace{0.5cm} \; index = min\_index(d)$
\STATE $\hspace{0.5cm} \; \theta[i] = index$
\ENDFOR
\end{algorithmic}
\label{alg2}
\end{algorithm}

\begin{figure*}[tp]
    \centering
    \includegraphics[width=1.0\linewidth]{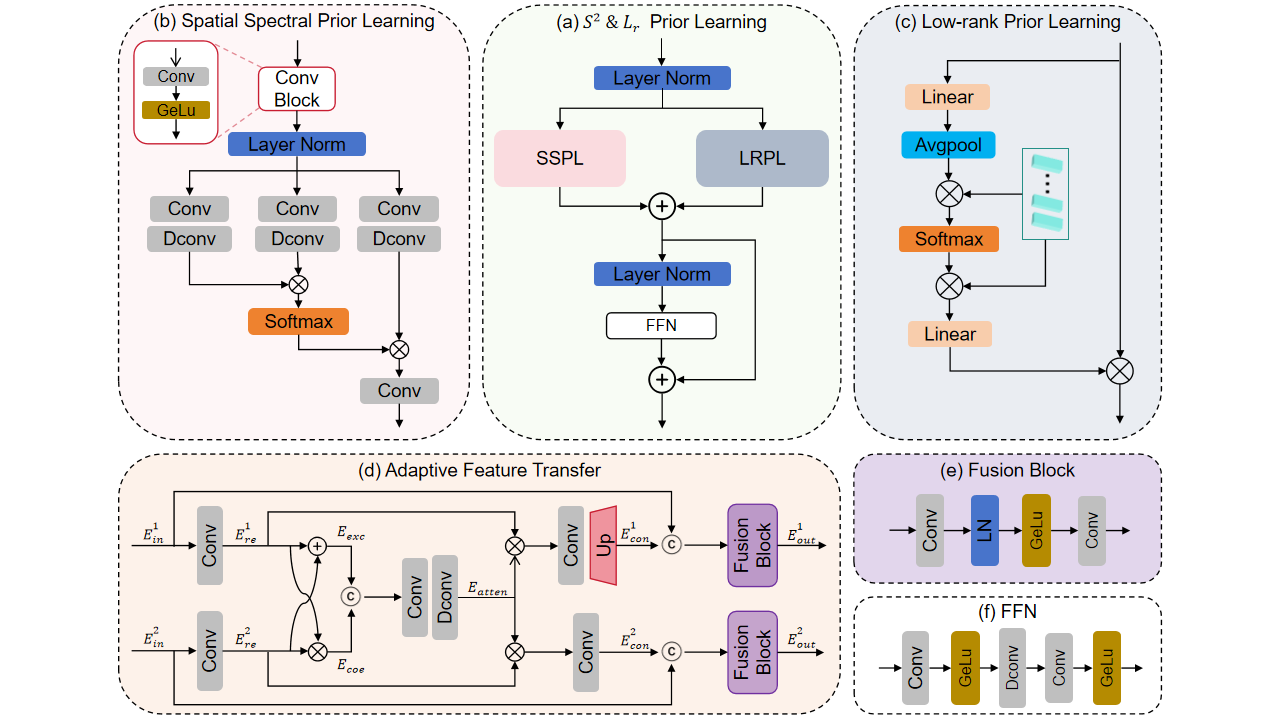}
    \caption{Diagram of the $S^{2} \& l_{r}$ Prior Learning. (a) The basic unit of the $S^{2} \& l_{r}$ Prior Learning Block; (b) The structure of the spatial-spectral prior learning branch; (c) The structure of the low-rank prior learning branch; (d) The structure of the adaptive feature transfer block; (e) The component of the Fusion block; (f) The component of the FFN network.}
    \label{fig3}
\end{figure*}

\subsection{Proximal mapping network}
\label{ssec:Proximal mapping network}
Fig. \ref{fig2} presents the  whole  architecture  of  our  proposed  ERRA  for  HSI reconstruction, which is composed of K stages to reconstruct a coded HSI. In each stage, a gradient descent module is followed by a proximal mapping module; The former aims to utilize transmittance information while the later is for optimization. Our proximal mapping module adopts a three-level U-shaped structure built using basic $S^{2}\& L_r$ prior learning block. The proximal mapping module uses a $conv \; 3 \times 3$ to map the output feature of upper layer $X_{k-1} \in R^{\; H\times W \times \lambda}$ into feature $X_{0}\in R^{\; H\times W \times C}$. $X_{0}$ passes through the encoder, bottleneck and decoder to be embedded into deep feature $X_{d}\in R^{\; H\times W \times C}$. Each level of the an encoder contains an $S^{2}\& L_r$ prior learning and a downsample module, while the decoder contains an upsample module and an $S^{2}$ prior learning module. Finally, a $\; 3 \times 3$ convolution operates on $X_{d}$ to generate output image $X_{k} \in R^{\; H\times W \times \lambda}$. We also introduce an adaptive feature transfer block, which dynamically transfers detailed and high-frequency features from the multi-scale encoder to the decoder. 

\textbf{$\bm{S^{2}}\boldsymbol{\&} \bm{L_r}$ Prior Learning Block} The $S^{2}\& L_r$ prior learning block plays an important role in realizing prior learning. It consists of two layer normalization steps, a spatial-spectral prior learning branch, a low-rank prior learning branch in a parallel design, and a feed-forward network, as shown in Fig. \ref{fig3} (f). The down-sampling module comprises a maxpool operation followed by a $3\times3$ conv layer. The up-sampling process involves a $3\times3$ ConvTransposed operation. Further details on the spatial-spectral prior learning branch and the low-rank prior learning branch will be provided subsequently.\par

\textbf{Spatial Spectral Prior Learning Branch.} A well designed proximal mapping module should adaptively learn HSI's local similarity and account for spectral nonlinear mapping. To achieve this, we propose SSPL to function as the feature extractor for both spatial and spectral dimensions. Fig. \ref{fig3}(b) shows the SSPL used in the first level. SSPL consists of a spatial CNN and spectral self-attention. The spatial part, with $conv 3 \times 3$ and $Gelu$, can effectively extract local contextual information. The spectral part computes cross-covariance across feature channels, exploring the global characteristics of HSI to gives more weight to importance features.\par

Spectral self-attention follows \cite{Restormer}. The key component of spectral self-attention is presented in Fig. \ref{fig3}(c). The input feature $X_{in} \in R^{\; H\times W \times C}$ is embedded as $Q,\;K,$ and $V$. Next, SSPL reshapes the query and key such that their dot-product interaction generates a transposed attention map $A\in R^{\; C \times C}$. The SSPL process is defined as:
{
\setlength\abovedisplayskip{10pt}
\setlength\belowdisplayskip{10pt}
\begin{align}
    & Attention(Q,K,V)=V\cdot Softmax(\frac{KQ}{\alpha}) \\
    & X_{SSPL} = W_{p}Attention(Q,K,V) 
\end{align}
}%
where $Q\in R^{\; HW \times C}$; $K\in R^{\; C \times HW}$; $V\in R^{\;HW \times C}$ are obtained after convolution and reshape operation from original size $\mathbb{R}^{\; H \times W \times C}$. $\alpha$ is a learnable scaling parameter to control the magnitude of the dot product $K$ and $Q$ before applying the softmax function.

\textbf{Low-rank Prior Learning Branch.} Low-rank representation of HSI can effectively maintain contextual relationships within high-dimensional structure. Several studies\cite{low-rank1, low-rank2, low-rank3, low-rank4} have demonstrated its effectiveness for HSI tasks. To explore the spectral low-rank property of HSIs, \cite{low-rank1} extended a deep CP Decomposition module to achieve low-rank prior learning. However, such a paradigm cannot learn optimal low-rank representation in deep unfolding network-based HSI reconstruction algorithms, since each deep prior network is independent, and the learned prior representation cannot be shared through stages. To address this challenge, we propose an inter-stages low-rank prior learning network. \par
The structure of low-rank prior learning module follows\cite{SENet}, but the difference is that we share low-rank prior information between several stages as illustrated in Fig. \ref{fig3}(c). Input features $X_{in} \in R \; ^{ H \times W \times C}$ are mapped into a 1D tensor $L_{c} \in R \; ^{1 \times 1 \times C} $ to aggregate global distribution in the spectral dimension. Then, it is projected into a subspace $L_{k} \in R \; ^{1 \times 1 \times C/r}$ of rank $C/r$ as follows:

{
\setlength\abovedisplayskip{10pt}
\setlength\belowdisplayskip{10pt}
\begin{align}
    & L_{c}=AveragePool(X_{in}) \\
    & L_{k}=Linear( L_{c})
\end{align}
}\par
To gather global spectral prior, a learnable query prior $Q_{k} \in R \; ^{m \times C/r}$ interact with the squeezed feature $L_{k}$, and the output of low-rank prior learning module is obtained by rescaling the HSI $X_{in} \in R \; ^{H \times W \times C}$ with informative low-rank attention $F_{atten} \in R \; ^{1 \times m}$ as:
{
\setlength\abovedisplayskip{10pt}
\setlength\belowdisplayskip{10pt}
\begin{align}
    & F_{atten}=Softmax(\frac{L_{k} Q_{k}^{T}}{\sqrt{C/r}}) \\
    & X_{LRPL} = X \cdot Linear(F_{atten}Q_{k})
\end{align}
}
where $\cdot$ denotes the element-wise dot product. In this module, the query prior function as an excitation to enhance the representation abilities of the squeezed feature, while the rescaling operation allows the current HSI to capture the low-rank prior of the entire HSI datasets. \par

\textbf{Adaptive Feature Transfer Block.} 
The encoder of U-Net acquires multi-scale detailed information from the image and gradually compresses the image size to capture global priors. The decoder then fuses up-sampled features with skip-connected features to progressively restore the image size. In this architecture, skip connections establish connections between the encoder and decoder, enabling the decoder to effectively utilize features from different levels and enhance the network's ability to preserve the details information. However, previous approaches that simply concatenate encoder features of the same scale\cite{ADMM-Net, DNUs3} failed to adequately transfer useful information, while roughly fusing multi-scales encoder features\cite{DNUs4} may introduce additional noise. Inspired by \cite{conv2former}, we propose 
an adaptive feature transfer block that models multi-scale contextual correlation features using convolution-style attention, and then adaptively transfers them to the decoder stage, adding only a small amount of computational overhead.\par

As presented in Fig. \ref{fig3}(d), the encoder feature $E_{1} \in R ^{H \times W \times C}$ and $E_{2}\in R ^{H \times W \times 2C}$ are processed through convolution operations to reduce the channel dimensionality and spatially down-sample $E_1$, and to only reduce the channel dimensionality of $E_2$:
{
\setlength\abovedisplayskip{10pt}
\setlength\belowdisplayskip{10pt}
\begin{align}
    & E_{re}^{1} = Conv(E_{in}^{1})\\
    & E_{re}^{2} = Conv(E_{in}^{2})
\end{align}
}%

After that, we utilize the product of the two-level  features $E_{exc} \in R^{\frac{H}{2} \times \frac{W}{2} \times \frac{C}{2}}$ to draw attention to important feature co-existing in both, sum of them $E_{coe} \in R^{\frac{H}{2} \times \frac{W}{2} \times \frac{C}{2}}$ to integrate their exclusive feature. To fully integrate $E_{exc}$ and $E_{coe}$, we use a $5\times 5$ depth-wise convolution on their concatenation to module contextual correlation attention as follows:
{
\setlength\abovedisplayskip{10pt}
\setlength\belowdisplayskip{10pt}
\begin{align}
    & E_{exc} = E_{re}^{1} \cdot E_{re}^{2}\\
    & E_{coe} = E_{re}^{1} + E_{re}^{2}\\
    & E_{atten} = DConv(E_{exc} \; \textcircled {c} \; E_{coe})
\end{align}
}%
where \textcircled {c} denotes the concatenation operation. Then, we further model multi-scales contextual correlation features with $E_{atten}$ and process them into the original scale:
{
\setlength\abovedisplayskip{10pt}
\setlength\belowdisplayskip{10pt}
\begin{align}
    & E_{con}^{1} = Up(Conv(E_{atten} \cdot E_{re}^{1}))\\
    & E_{con}^{2} = Conv(E_{atten} \cdot E_{re}^{2})
\end{align}
}%
where $E_{con}^{1} \in R^{H \times W \times C}$,  $E_{con}^{2} \in R^{\frac{H}{2} \times \frac{W}{2}  \times 2C}$ are processed contextual correlation features. Finally, we fuse them with the origin encoder output to transfer enriched detailed feature to the decoder:
{
\setlength\abovedisplayskip{10pt}
\setlength\belowdisplayskip{10pt}
\begin{align}
    & E_{out}^{1} = Fusion(E_{in}^{1} \; \textcircled {c} \; E_{con}^{1})\\
    & E_{out}^{2} = Fusion(E_{in}^{2} \; \textcircled {c} \; E_{con}^{2})
\end{align}
}%
where $Fusion$ denotes the feature fusion block as presented in Fig. \ref{fig3}(e). The adaptive feature transfer network can collectively model multi-scale encoder features, thereby preventing the information loss when features are transferred level by level as in previous works.

\begin{table*}[!t]
\caption{Comparison of results on the AVIRIS-NG dataset, where the top entry in each cell is the PSNR metric in dB and the bottom entry in each cell is the SSIM metric. The best results are in {\color{red} red} and the second-best results are in {\color{blue} blue}.}
\label{tab:table11}
\centering
\tabcolsep=3pt
\renewcommand\arraystretch{1.5}
\begin{tabular}{cccccccccccccc}
\Xhline{1.5pt}
\textbf{Methods} & \textbf{Scene01} & \textbf{Scene02} & \textbf{Scene03} & \textbf{Scene04} & \textbf{Scene05} & \textbf{Scene06} & \textbf{Scene07} & \textbf{Scene08} & \textbf{Scene09} & \textbf{Scene10} & \textbf{Scene11} & \textbf{Scene12} &\textbf{average}\\

\hline
\multirow{2}{*}{TSANet\cite{meng2020end}}
& 42.08 & 42.86 & 43.06 & 36.95 & 38.31 & 37.02 & 38.10 & 40.85 & 39.16 & 39.53 & 39.64 & 39.33 & 39.74\\ 
& 0.990 & 0.992 & 0.993 & 0.971 & 0.975 & 0.970 & 0.981 & 0.990 & 0.987 & 0.985 & 0.986 & 0.984 & 0.984 \\ 
\hline
\multirow{2}{*}{ADMM-Net\cite{ADMM-Net}}
& 42.26 & 43.63 & 43.81 & 37.65 & 38.84 & 37.77 & 37.87 & 40.73 & 39.35 & 39.56 & 39.58 & 39.36 & 40.03\\ 
& 0.990 & 0.993 & 0.994 & 0.975 & 0.978 & 0.973 & 0.979 & 0.991 & 0.988 & 0.985 & 0.986 & 0.985 & 0.985 \\ 
\hline
\multirow{2}{*}{GAP-Net\cite{DNUs3}}
& 42.30 & 43.74 & 43.91 & 37.60 & 38.87 & 37.85 & 37.85 & 40.78 & 39.46 & 39.52 & 39.52 & 39.27 & 40.06 \\ 
& 0.990 & 0.993 & 0.994 & 0.975 & 0.978 & 0.973 & 0.979 & 0.991 & 0.988 & 0.985 & 0.986 & 0.984 & 0.985\\ 
\hline
\multirow{2}{*}{HDNet\cite{hdnet}}
& 42.71 & 44.50 & 44.49 & 39.35 & 40.33 & {\color{blue} 39.69} & 39.26 & 41.77 & 39.97 & 40.12 & 40.20 & 40.21 & 41.05 \\ 
& 0.991 & 0.995 & 0.995 & 0.982 & 0.984 & {\color{blue} 0.982} & 0.984 & 0.992 & 0.989 & 0.987 & 0.988 & 0.988 & 0.988\\ 
\hline
\multirow{2}{*}{RDLUF\cite{DNUs4}}
& 42.95 & 44.65 & 44.66 & {\color{red} 39.62} & {\color{red} 40.41} & 39.65 & {\color{blue} 39.77} & 42.08 & {\color{blue} 40.45} & {\color{red} 40.81} & 40.49 & 40.43 & {\color{blue} 41.33} \\ 
& 0.991 & 0.994 & 0.994 & {\color{red} 0.983} & {\color{red} 0.984} & 0.980 & {\color{blue} 0.985} & 0.991 & 0.989 & {\color{red} 0.988} & 0.989 & 0.988 & {\color{blue} 0.988}\\
\hline
\multirow{2}{*}{Ours 3stage}
& 38.12 & 40.67 & 39.99 & 36.88 & 37.37 & 33.16 & 37.22 & 38.37 & 38.15 & 38.75 & 38.93 & 39.10 & 38.06 \\ 
& 0.984 & 0.990 & 0.989 & 0.976 & 0.977 & 0.970 & 0.980 & 0.986 & 0.986 & 0.984 & 0.986 & 0.985 & 0.983\\ 
\hline
\multirow{2}{*}{Ours 5stage}
& 39.73 & 42.82 & 42.05 & 37.81 & 38.32 & 34.83 & 37.80 & 39.42 & 38.66 & 39.39 & 39.54 & 39.58 & 39.16 \\ 
& 0.984 & 0.991 & 0.990 & 0.978 & 0.978 & 0.970 & 0.980 & 0.986 & 0.986 & 0.984 & 0.986 & 0.986 & 0.983\\ 
\hline
\multirow{2}{*}{Ours 7stage}
& {\color{blue} 43.23} & {\color{blue} 44.80} & {\color{blue} 44.71} & 39.06 & 39.99 & 39.04 & 39.27 & {\color{blue} 42.17} & 40.23 & 40.53 & {\color{blue} 40.68} & {\color{blue} 40.64} & 41.20 \\ 
& {\color{blue} 0.992} & {\color{blue} 0.995} & {\color{blue} 0.994} & 0.981 & 0.983 & 0.979 & 0.984 & {\color{blue} 0.992} & {\color{blue} 0.990} & 0.988 & {\color{blue} 0.989} & {\color{blue} 0.988} & 0.988\\ 
\hline
\multirow{2}{*}{Ours 9stage}
& {\color{red} 43.43} & {\color{red} 45.09} & {\color{red} 45.06} & {\color{blue} 39.54} & {\color{blue} 40.35} & {\color{red} 39.91} & {\color{red} 40.17} & {\color{red} 42.51} & {\color{red} 40.60} & {\color{blue} 40.64} & {\color{red} 40.75} & {\color{red} 40.79}& {\color{red} 41.57} \\ 
& {\color{red} 0.992} & {\color{red} 0.995} & {\color{red} 0.995} & {\color{blue} 0.983} & {\color{blue} 0.984} & {\color{red} 0.982} & {\color{red} 0.987} & {\color{red} 0.992} & {\color{red} 0.990} &{\color{blue} 0.988} & {\color{red} 0.990} & {\color{red} 0.989} & {\color{red} 0.989}\\ 
\hline
\end{tabular}
\end{table*}
\begin{figure*}[h!]
    \centering
    \includegraphics[width=1.0\linewidth]{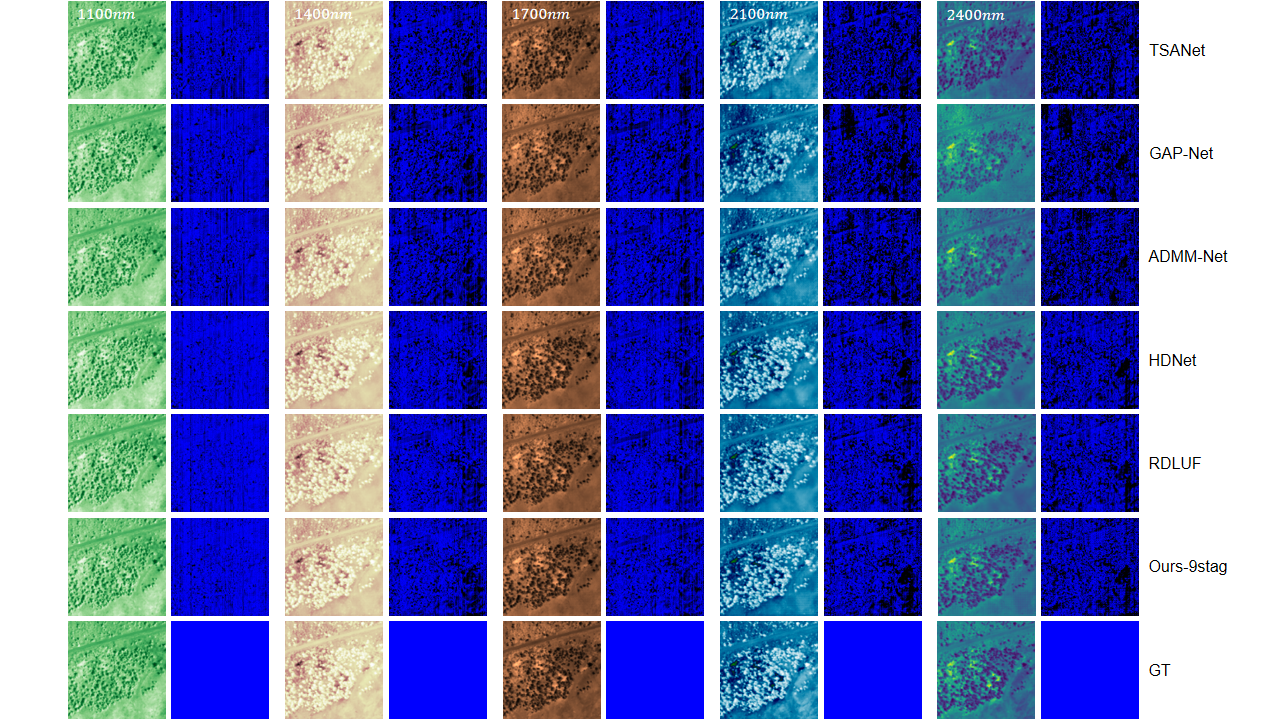}
    \caption{The simulation reconstruction results of Scene03 in the test set. The results are compared with the other five methods in five different spectral bands. Residual maps are shown next to each reconstruction to facilitate comparison.}
    \label{fig4}
\end{figure*}

\begin{figure*}[h]
    \centering
    \includegraphics[width=1.0\linewidth]{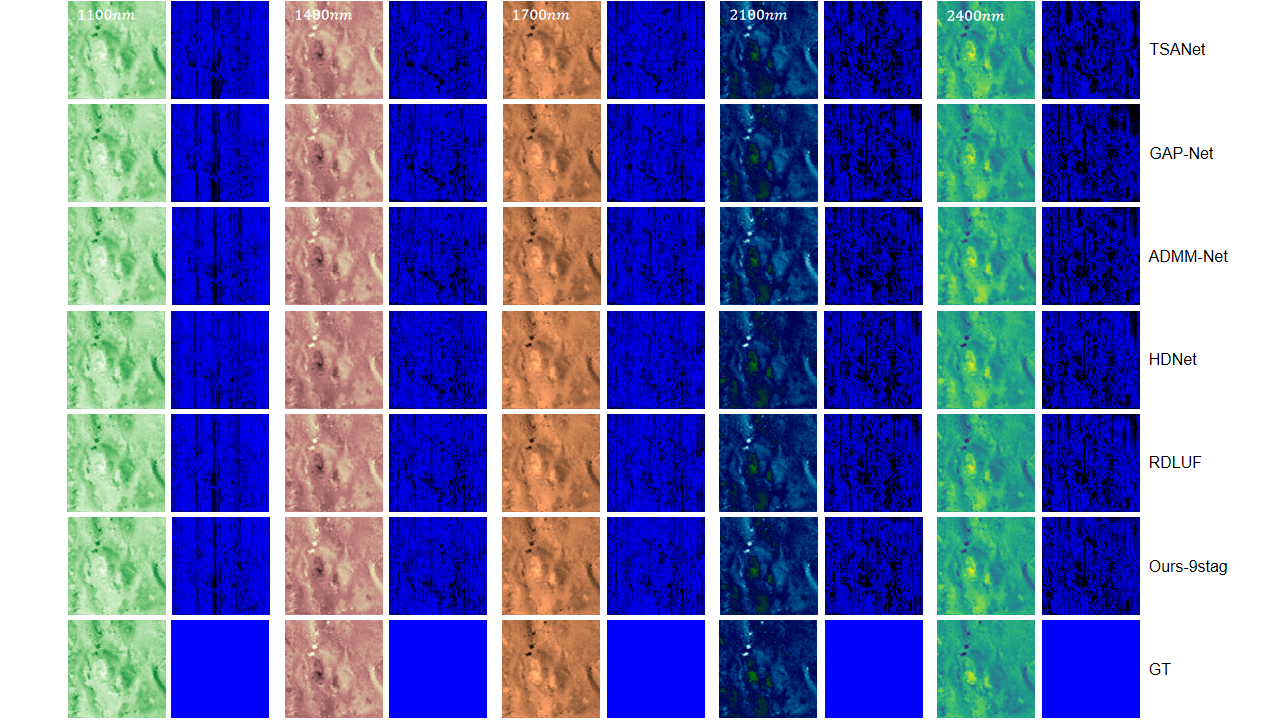}
    \caption{The simulation reconstruction results of Scene05 in the test set. The results are compared with the other five methods in five different spectral bands. Residual maps are shown next to each reconstruction to facilitate comparison.}
    \label{fig5}
\end{figure*}

\begin{figure*}[h]
    \centering
    \includegraphics[width=1.0\linewidth]{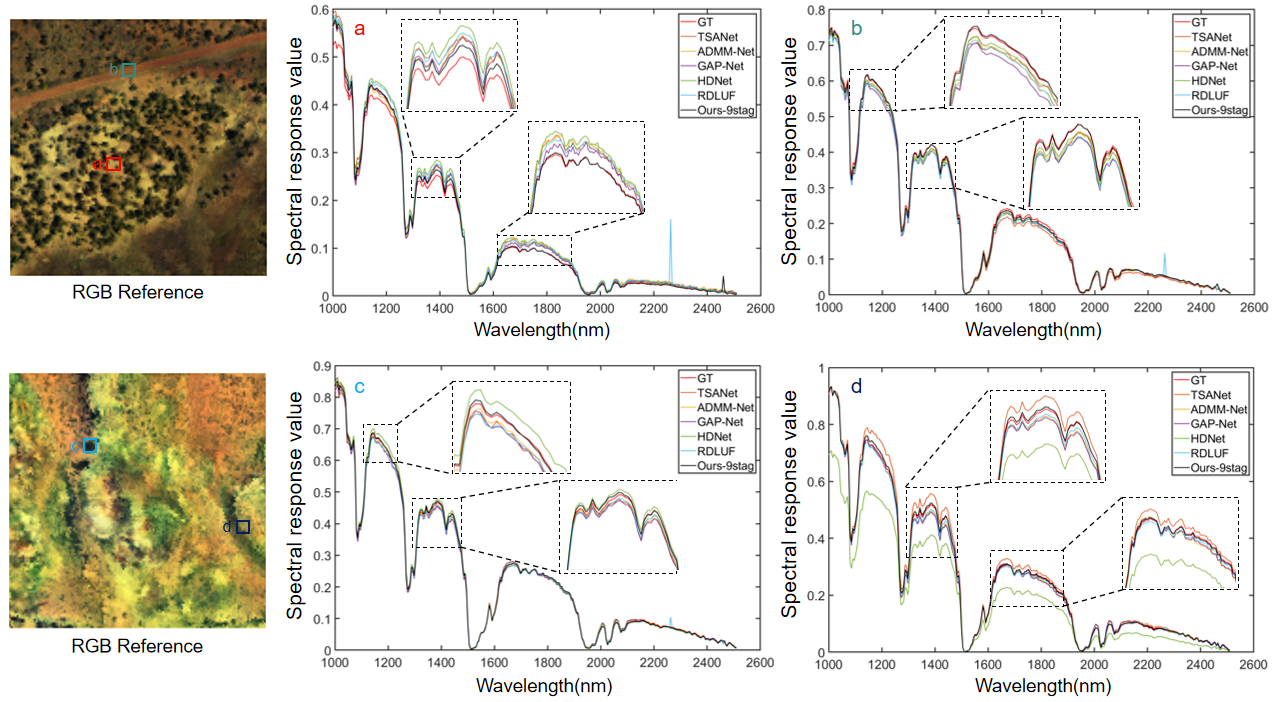}
    \caption{Spectral curve comparison of different reconstruction results. Four areas (a, b, c, and d) are selected on the RGB data of scene03 and scene05, and the spectral curves of these areas. The horizontal axis represents different spectral bands, and the vertical axis represents the average intensity value of different regions across these spectral bands.}
    \label{fig6}
\end{figure*}

\section{Experiments}
\label{sec:Experiments}
In this section, we compare our method with several state-of-the-art (SOTA) methods on the AVIRIS-NG (Airborne Visible InfraRed Imaging Spectrometer Next Generation) dataset. We selected 300 wavelengths ranging from 1000nm to 2500nm and removed ands whose response values were zero due to water vapor absorption. The peak-signal-to-noise-ratio (PSNR) and structured similarity index (SSIM) metrics are used to evaluate the performance of different hyperspectral image reconstruction methods.
\subsection{Datasets}
AVIRIS-NG measures wavelengths ranging from 380nm to 2510nm with a 5nm sampling interval. Spectra are captured as images with 600 cross-track elements and spatial sampling ranging from 0.3m to 4.0m. We selected 12 flight lines, and each hyperspectral image is approximately sized at 21K $\times$ 0.65K $\times$ 432. We create spatial domain tiles for each image, with each tile sized at 256 $ \times $ 256 $\times$ 300.

\begin{figure*}[h]
    \centering
    \includegraphics[width=1.0\linewidth]{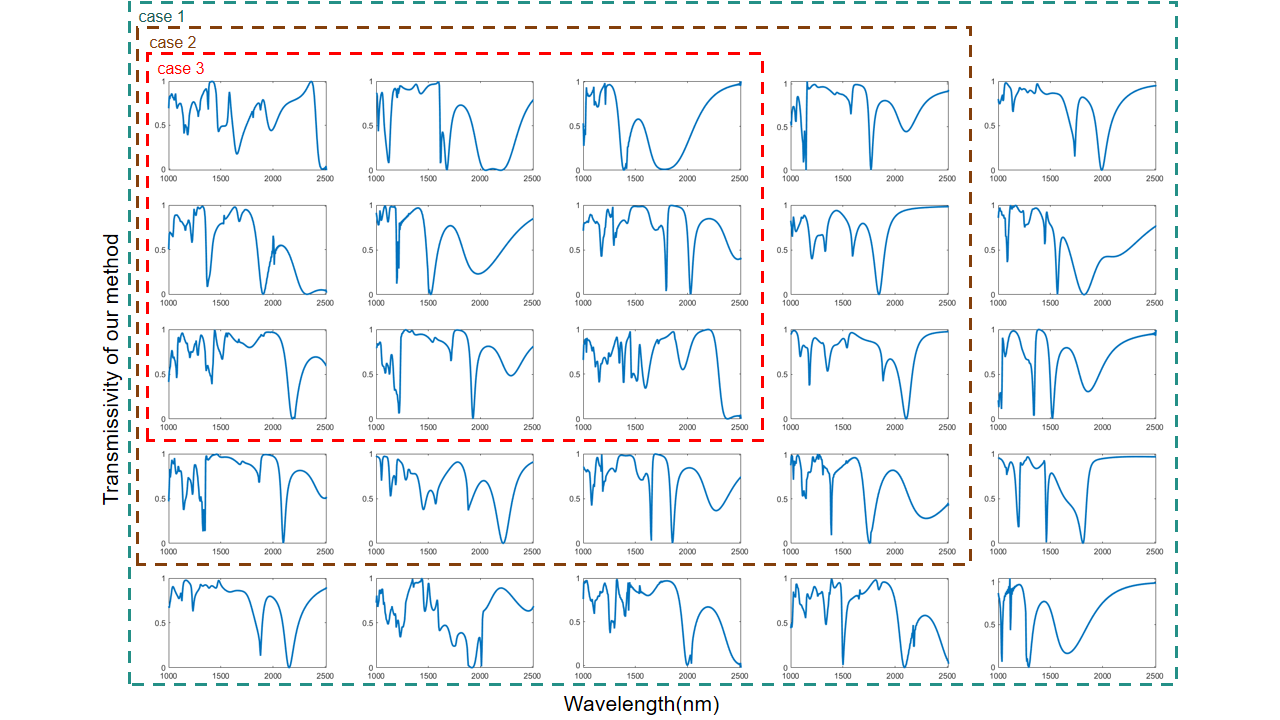}
    \caption{The transmissivity of the metasurface filter selected by our proposed filter selection method. Case 1: basic filter array with a 5$\times$5 layout; Case 2: basic filter array with a 4$\times$4 layout; Case 3: basic filter array with a 3$\times$3 layout;}
    \label{fig7}
\end{figure*}

\subsection{Implementation Details}
Our model was implemented using the PyTorch framework and trained with the Adam optimizer for $300$ epochs. During the training, we applied random cropping,  rotation, and flipping of tiles for data augmentation. The learning rate was set to 0.0002 and the batch size 1. The training and test data size was 64 $\times$ 64 $\times$ 300. All experiments were conducted on the NVIDIA Tesla P40 GPU.
\subsection{Evaluation Indexes}
We adopt two image quality indexes, including peak signal-to-noise ratio (PSNR) and structure similarity (SSIM)\cite{SSIM}, for quantitative evaluation. Specifically, PSNR measures the visual quality, SSIM measures the structure similarity. Generally, bigger values of PSNR and SSIM suggest a better performance.

\subsection{Comparisons with SOTA Methods}
We conducted a comprehensive comparative analysis of the proposed method and various SOTA methods, including two end-to-end methods (TSANet\cite{meng2020end},HDNet\cite{hdnet}) and three deep unfolding networks (GAP-Net\cite{DNUs3},ADMM-Net\cite{ADMM-Net},RDLUF\cite{DNUs4}). It is worth noting that RDLUF is implemented with help of the cross-stages interaction and multi-scales encoder feature fusion. We replaced the CASSI simulation imaging part of all comparison methods with the simulation process of filter array imaging, adjusting the number of feature channels, and maintaining consistency in learning rate, loss function, optimizer settings as much as possible. \par

Table \ref{tab:table11} presents 12 scenes and average results of PSNR and SSIM for the different comparison methods. Our method achieve best performance with both 41.57dB PSNR and 0.989 SSIM. TSANet learns both spatial and spectral correlations but shows the worst performance. Relative to the best counterpart method (RDLUF), another inter-stages feature learning method, our method improves by an average of 0.24dB improvement in PSNR and a 0.001 improvement in SSIM. Different deep unfolding models (GAP-Net,ADMM-Net) simply uses a U-Net to learn the deep prior, without considering the inherent characteristics of HSI, leading to limited performance. HDNet presents results that are slightly lower than our method. Moreover, we find that model based method like DeSCI\cite{model2}, GAP-TV\cite{GAP} cannot resolve HSI reconstruction task on AVIRIS-NG dataset, we think that excessive number of bands and severe spectral variation result in this situation. In conclusion, the significant improvements show that the proposed method is effective for HSI reconstruction. \par

To compare our method with other state-of-the-art methods qualitatively, we have selected several methods and showcased the results in Scene03 across five spectral bands in Fig. \ref{fig4}. The reconstruction results of our method and five others are presented, along with the respective residual maps for each reconstruction. Notably, our method demonstrates lower reconstruction errors, especially in the top right corner. Furthermore, in a grassland texture-based Scene05, as depicted in Fig. \ref{fig5}, our method closely aligns with the ground truth in terms of visual effect, while other methods exhibit fewer details and more artifacts in fine features.

\begin{table}[H]
\caption{Computational complexity and average reconstruction quality of several SOTA methods on the AVIRIS-NG dataset}
\label{tab:table2}
\centering
\tabcolsep=6pt
\renewcommand\arraystretch{2}
\begin{tabular}{ccccc}
\toprule[1pt]
\textbf{Method} & \textbf{Params(M)} & \textbf{FLOPs(G)} & \textbf{PSNR(dB)} & \textbf{SSIM}\\
\midrule
TSANet & 162.87 & 361.37 & 39.74 & 0.984\\
ADMM-Net & 266.37 & 331.24 & 40.03 & 0.985\\
GAP-Net & 266.37 & 331.24 & 40.06 & 0.985\\
HDNet & 529.64 & 2167.27 & 41.05 &0.988\\
RDLUF & 289.63 & 974.577 & 41.33& 0.988\\
Ours-9stag & 517.46 & 1561.29 & 41.57 & 0.989\\
\bottomrule[1pt]
\end{tabular}
\end{table} 

For a more detailed quantitative analysis of the reconstructions in local regions, we illustrated two selected regions from Scene03 and Scene05 in Fig. \ref{fig6}, labeled as a, b, c, and d. We presented the spectral curves and computed their correlation with the ground truth data, comparing the results with five state-of-the-art methods. Our method's spectral curve is represented by the black curve, closely resembling the red curve of the ground truth data. \par
In a comprehensive comparison, we analyzed the complexity of our proposed algorithm alongside other state-of-the-art reconstruction algorithms, as displayed in Table \ref{tab:table2}. While our method demonstrates a moderate level of computational complexity, it requires significantly more floating-point operations compared to the deep unfolding method RDLUF \cite{DNUs4}.

\subsection{Ablation Study}
We conducted ablation study to  evaluate  the  effectiveness  of  various  components  of our proposed method. \par

\textbf{Break-down Ablaton.} To assess the efficacy of the different components in our approach, we conducted a series of ablation experiments with k set to 9 stages. Table \ref{tab:table3} presents the PSNR and SSIM results for each component. Initially, we substituted the complete ERRA network with a spatial-spectral prior learning architecture as the base framework, as shown in row (a) of Table \ref{tab:table3}. The LRPL module demonstrated an increase of 0.67dB in PSNR compared to the base framework, showcasing the effectiveness of our cross-stage low rank prior learning method, as seen in row (b) of Table \ref{tab:table3}. Additionally, we integrated the Adaptive Feature Transfer (AFT) module into the LRPL framework to assess the interaction of feature information. The results in row (c) of Table \ref{tab:table3} indicate that the AFT module achieved a 0.44dB higher PSNR compared to the LRPL framework, confirming its efficacy across different stages of the network.

\textbf{Number of stages.} 
We conducted experiments at various stages to analyze the impact of the unfolding network on overall expressiveness. The results are presented in Table \ref{tab:table11}, showcasing the computational complexity and average reconstruction quality on the AVIRIS-NG test dataset. With an increasing number of network stages, there is a gradual rise in floating-point calculations, resulting in improved experimental outcomes. To find a balance between impact and complexity, we settled on 9 as the final number of stages for the network.

\begin{table}[H]
\caption{The effectiveness of different components}
\label{tab:table3}
\centering
\renewcommand\arraystretch{1.5}
\begin{tabular}{cccccc}
\toprule[1pt]
\textbf{setting} & \textbf{SSPL} & \textbf{LRPL} & \textbf{AFT} & \textbf{PSNR(dB)} & \textbf{SSIM}\\
\midrule
a & \checkmark &  &  & 40.46 & 0.984\\

b & \checkmark & \checkmark & & 41.13 & 0.988\\

c & \checkmark & \checkmark & \checkmark & 41.57 & 0.989\\
\bottomrule[1pt]
\end{tabular}
\end{table}

\begin{figure}[H]
    \centering
    \includegraphics[width=0.8\linewidth]{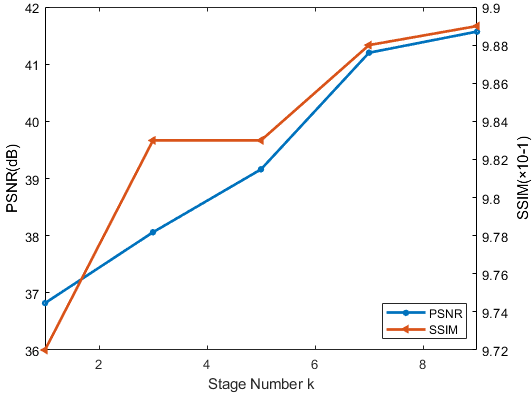}
    \caption{PSNR and SSIM curves with the increasing number of stages k.}
    \label{fig8}
\end{figure}

Moreover, as the network progresses through more stages, there are minimal parameter changes due to our strategy of parameter sharing across stages during training. Additionally, a line graph in Fig. \ref{fig8} visually represents the stage changes and the network's evolution. The horizontal axis denotes the number of stages (k) in the unfolding network, the left axis shows the PSNR results, and the right axis displays the SSIM values. Notably, as the stage number increases, both PSNR and SSIM metrics show progressive enhancement.

\subsection{Pattern layout analysis}
To verify the importance of the pattern layout in enhancing the reconstruction quality, we conducted a comparative experiment using different filter selection methods and varying filter numbers. \par
\textbf{Number of metasurface filter.} For the optimization of the basic filter array, we replaced the basic filter array with different number of filters(i.e, Case 1, Case 2, Case 3) as shown in Fig. \ref{fig7}, and all of the basic filter arrays were generated using our proposed filter selection method. The experimental results indicate that a smaller number of filters leads to better reconstruction performance as presented in Table \ref{tab:table4}. Since the spectral information required for reconstructing the spectrum of each pixel is captured by various filters within the local region, a large number of filters in basic array may introduce incorrect spectral information that differs from the central pixel due to the significant variations in spectral features of remote sensing images. This can adversely affect spectral reconstruction and result in a decrease in reconstruction performance. Hence, we have chosen Case 3 as our standard basic filter array.

\begin{table}[H]
\caption{Performance with different number of filters}
\label{tab:table4}
\centering
\renewcommand\arraystretch{1.5}
\begin{tabular}{ccccc}
\toprule[1pt]
\textbf{Metric} & \textbf{Case 1} & \textbf{Case 2} & \textbf{Case 3} & \textbf{Case 4}\\
\midrule
PSNR(dB) & 40.17 & 40.41 & 41.20 & 40.57\\

SSIM & 0.984 & 0.985 & 0.988 & 0.983\\
\bottomrule[1pt]
\end{tabular}
\end{table}

\begin{figure}
    \centering
    \includegraphics[width=1\linewidth]{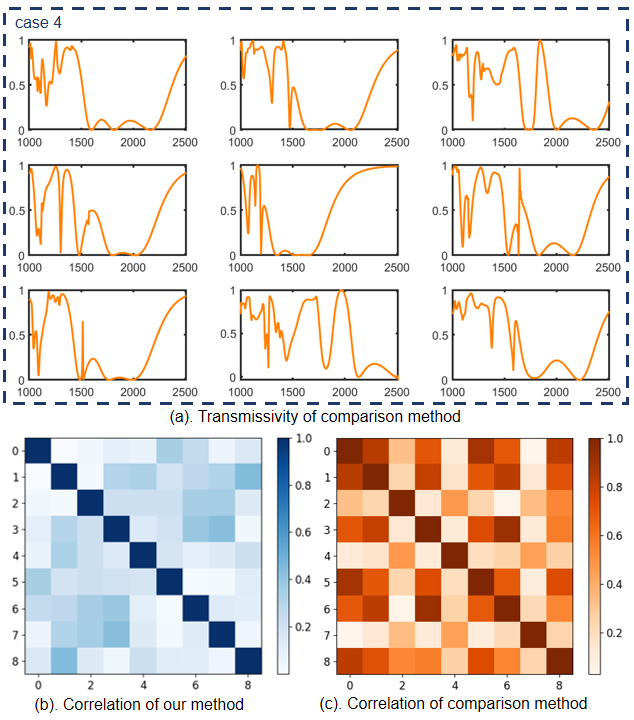}
    \caption{(a) The transmissivity of the metasurface filter selected by the existing method; (b) The correlation between filters in the array selected by our filter selection method; (c) The correlation between filters in the array selected by the existing method.}
    \label{fig9}
\end{figure}

\textbf{Filter selection method.} In order to verify the effectiveness of the our proposed filter selection method combined with the farthest distance sampling and Pearson correlation coefficient, we conducted experiments under the filter array selected by two methods (inner product-based optimization method\cite{dynamic_brain} and our method), the transmissivity of existing method is shown in Fig. \ref{fig9}(a)(Case 4). As presented in Table\ref{tab:table4}, existing filter selection method(Case 4) shows a large lower than our method(Case 3) by 0.63dB in PSNR and 0.005 in SSIM. Besides, the visualization results also show that the randomness of metasurface filter selected by our method better than existing method, and the correlation between the metasurface units in the basic filter array changes more smoothly as shown in Fig. \ref{fig9}(b-c).

\section{Conclusion}
\label{sec:Conclusion}
In this paper, we enhance the correlation coefficient based filter selection and propose an inter and intra prior learning deep unfolding network for SWIR-HSI reconstruction. By intefrating FPS into the original filter selection, we achieve better optimization of filter arrays and higher reconstruction quality. For integrating low rank prior learning and cross-stage information interaction, a cross-stage low rank prior learning module is proposed to globally capture HSI's useful characteristics. To address the loss of detailed features in the network, a adaptive feature transfer module is introduced to adaptively fuse the detail information from encoder stages and transfer them to decoder stages.  Experimental results demonstrate that our method outperforms previous HSI reconstruction network.\\

\bibliographystyle{IEEEtran}
\bibliography{reference.bib}

\end{document}